\documentclass{article}

\usepackage{verbatim}
\usepackage{listings}

\usepackage{amsmath}
\usepackage{algorithm}
\usepackage{subcaption}
\usepackage{graphicx}
\usepackage[noend]{algpseudocode}
\usepackage{theorem}

\usepackage[utf8]{inputenc}

\theoremstyle{plain}
\newtheorem{exmp}{Exemple}[section]
\newtheorem{mydef}{D{\'e}finition}[section]

\date{Juin 2016}
\title{Une approche totalement instanciée pour \\la planification HTN}

\author{Abdeldjalil Ramoul$^{1, 2}$, Damien Pellier$^{1}$, \\
Humbert Fiorino$^{1}$ and Sylvie Pesty$^{1}$ \\[2mm]
$^{1}$Université Grenoble Alpes\\
Laboratoire d’Informatique de Grenoble\\
700 avenue Centrale - 38401 - Saint-Martin-d’Hères, France\\[1mm]
$^{2}$Intrinsec\\
215 Avenue Georges Clemenceau - 92024 - Nanterre, France\\
}

\begin{document}
\maketitle

\begin{abstract}
De nombreuses techniques de planification ont été développées pour permettre à des systèmes autonomes d'agir et de prendre des décisions en fonction de leurs perceptions de l'environnement. Parmi ces techniques, la planification HTN ({\it Hierarchical Task Network}) est l'une des techniques les plus utilisées en pratique. Contrairement aux approches classiques de la planification, la planification HTN fonctionne par décomposition récursive d'une tâche complexe en sous tâches jusqu'à ce que chaque sous-tâche puisse être réalisée par l'exécution d'une action. Cette vision hiérarchique de la planification permet une représentation plus riche des problèmes de planification tout en guidant la recherche d'un plan solution et en apportant de la connaissance aux algorithmes sous-jacents. Dans cet article, nous proposons une nouvelle approche de la planification HTN dans laquelle, comme en planification classique, nous instancions l'ensemble des opérateurs de planification avant d'effectuer la recherche d'un plan solution. Cette approche a fait ses preuves en planification classique. Elle est utilisée par la plupart des planificateurs contemporains mais n'a, à notre connaissance, jamais été appliquée dans le cadre de la planification HTN. L'instanciation des opérateurs de planification est pourtant nécessaire au développement d'heuristiques efficaces et à l'encodage de problèmes de planification HTN dans d'autres formalismes tels que SAT ou CSP. Nous présentons dans la suite de l'article un mécanisme générique d'instanciation. Ce mécanisme implémente des techniques de simplification permettant de réduire la complexité du processus d'instanciation inspirées de celles utilisées en planification classique. Pour finir nous présentons des résultats obtenus sur un ensemble de problèmes issus des compétitions internationales de planification avec une version modifiée du planificateur SHOP utilisant notre technique d'instanciation.
\end{abstract}

%---------------------------------------------------------------
%		Section 1: Introduction
%---------------------------------------------------------------

\section{Introduction}

% Contexte générale
Agir et prendre des décisions rationnelles en fonction des perceptions de l'environnement est une problématique centrale des systèmes autonomes et intelligents. La planification en cherchant à rendre calculable ce processus de décision a développé de nombreuses techniques pour répondre à cette problématique. Parmi l'ensemble de ces techniques, la planification HTN ({\it Hierarchical Task Network}) est l'une des techniques qui est la plus utilisée en pratique \cite{weser2010HTNrobotics,bevacqua2015HTNdrones,strenzke2011HTNaircrafts} pour des raisons d'efficacité mais aussi pour l'expressivité des langages HTN qui permettent de spécifier dans les domaines de planification des connaissances métier de haut niveau d'abstraction pouvant être utilisées par les algorithmes sous-jacents pour guider de manière efficace la recherche d'un plan solution. Contrairement à la planification classique \cite{fikes71strips} où le but est défini comme un ensemble de propositions à atteindre, en planification HTN le but s'exprime comme une tâche ou un ensemble de tâches à réaliser auxquelles il est possible d'associer des contraintes. Ce couple tâches contraintes est appelé un {\it réseau de tâches}. La recherche d'un plan solution consiste à décomposer le réseau de tâches initiales définissant le but, en respectant les contraintes spécifiées, en un ensemble de sous-tâches primitives qui peuvent être exécutées par une action au sens classique de la planification. La décomposition est réalisée en appliquant des règles de décomposition définies par des opérateurs hiérarchiques de planification appelés {\it méthodes}. Chaque méthode définit une décomposition possible d'une tâche en un ensemble de sous-tâches avec les contraintes qui les lient. La décomposition se termine lorsque le processus de décomposition aboutit à un réseau de tâches contenant uniquement des tâches primitives exécutables par une action et dont l'ensemble des contraintes associées sont vérifiées.

% État de l'art HTN
Les planificateurs HTN peuvent être divisés en deux grandes familles \cite{georgievski15}. Cette division s'appuie sur la nature de l'espace de recherche utilisé par ces algorithmes: recherche dans un espace de plans ou dans un espace d'états. Dans la première catégorie, l'espace de recherche est constitué de plans sous forme de réseau de tâches dans lequel les planificateurs ne maintiennent pas d'état pendant la recherche. À chaque étape du processus de recherche le réseau de tâches obtenu après une décomposition est considérée comme un plan partiel avec des contraintes qu'il faut satisfaire dans les décompositions suivantes. Cette représentation de l'espace de recherche permet de décomposer le réseau initial de tâches en un plan solution partiellement ordonné. On peut citer dans cette catégorie, les planificateurs \textit{NOAH (Nets Of Action Hierarchies)} \cite{sacerdoti75, sacerdoti75nl}, \textit{Nonlin (Non-Linear Planner)} \cite{tate76, tate77} , \textit{O-Plan} \cite{currie91_O-plan} et \textit{O-Plan2} \cite{tate94_o-plan2}, \textit{SIPE (System for Interactive Planning and Execution)} \cite{wilkins84SIPE} et \textit{SIPE-2} \cite{wilkins90SIPE-2}. En 1994 l'algorithme \textit{UMCP (Universal Method Composition Planner)}\cite{umcp94} est le premier algorithme HTN dont la correction et la complétude sont prouvées.
Dans la seconde catégorie, les planificateurs maintiennent des états pendant la recherche. On associe ces états aux réseaux de tâches. Le processus de décomposition commence en choisissant les tâches à décomposer dans l'ordre d'exécution, puis applique la méthode de décomposition dont les préconditions sont vérifiées dans l'état précédant la tâche décomposée. On peut citer par ordre chronologique comme planificateur de cette catégorie, \textit{SHOP} \cite{shop99}, \textit{SHOP2} \cite{shop203} et \textit{SIADEX} \cite{de2005siadex}.

% Problème abordé dans le papier
Parallèlement au développement de la planification HTN, de nombreux algorithmes de planification performants n'utilisant pas de représentation hiérarchique ont été développés, par exemple \textit{Fast Forward} \cite{hoffmann01ff} ou encore \textit{Fast Downward} \cite{helmert2006fast,seipp14cedalion}. Ils reposent tous sur une étape de pré-traitement qui consiste à dénombrer les actions possibles à partir des opérateurs de planification définis dans le domaine. Cette étape est cruciale pour ces algorithmes pour plusieurs raisons. Tout d'abord, cette étape de dénombrement ou d'instanciation permet de réduire le nombre d'actions du problème de planification grâce à des mécanismes de simplification. Ceci a pour conséquence de réduire le coefficient de branchement de l'espace de recherche. En second lieu, le fait de dénombrer l'ensemble des actions d'un problème de planification permet de réaliser une étude a priori des propriétés du monde atteignables. Cette étude est un préalable indispensable à l'élaboration et au développement d'heuristiques efficaces pour guider la recherche d'un plan solution \cite{hoffmann01ff,geffner2000heur,haslum2005newheur,hoffmann2004landmarks,richter2008landmarks,helmert2014merge}. Troisièmement, cette étape de prétraitement est un prérequis nécessaire à l'encodage d'un problème de planification dans des formalismes tels que CSP \cite{kambhampati2000CSP,bartak2010CSP,lopez2003CSP} ou SAT \cite{kautz1992sat,rintanen2012sat,rintanen14madagascar,kautz1999sat}. Toutefois, à notre connaissance, ce prétraitement n'a jamais été réalisé et adapté dans un contexte HTN. Pour toutes ces raisons, il serait intéressant d'effectuer l'instanciation et une simplification des opérateurs de planification, en intégrant la dimension hiérarchique, afin de réduire le nombre de décompositions possibles et la complexité de l'espace de recherche.

% Contribution
Dans cet article nous proposons une extension des mécanismes d'instanciation classiques pour la planification dans un contexte HTN. Nous montrons notamment comment il est possible de généraliser les mécanismes de simplification développés pour l'instanciation des opérateurs classiques de planification à l'instanciation des méthodes. Pour cela, nous réutilisons les règles d'instanciation et de simplification des opérateurs et des prédicats développées en planification non hiérarchique et définissons de nouvelles règles pour les méthodes de décomposition adaptées aux problèmes de planification HTN.

% Énoncé du plan
Dans un premier temps, nous commençons par donner un exemple de problème HTN qui sert de fil conducteur tout au long de l'article et présentons le formalisme HTN. Par la suite nous introduisons le  problème de l'instanciation des problèmes de planification HTN et explicitons sa complexité. Dans un second temps, nous détaillons les mécanismes d'instanciation et de simplification proposés pour les problèmes de planification HTN et présentons une version modifiée de l'algorithme SHOP que nous avons nommé \textit{iSHOP (instantiated SHOP)} qui prend en entrée un problème totalement instancié. Enfin, nous montrons dans quelle mesure notre version de SHOP améliore les performances en termes de temps de recherche d'un plan solution par rapport à l'algorithme SHOP.

%---------------------------------------------------------------
%		Section 2: Un exemple introductif
%---------------------------------------------------------------

\section{Un exemple introductif}
\begin{figure}[h]
\centerline{\includegraphics[scale=0.42]{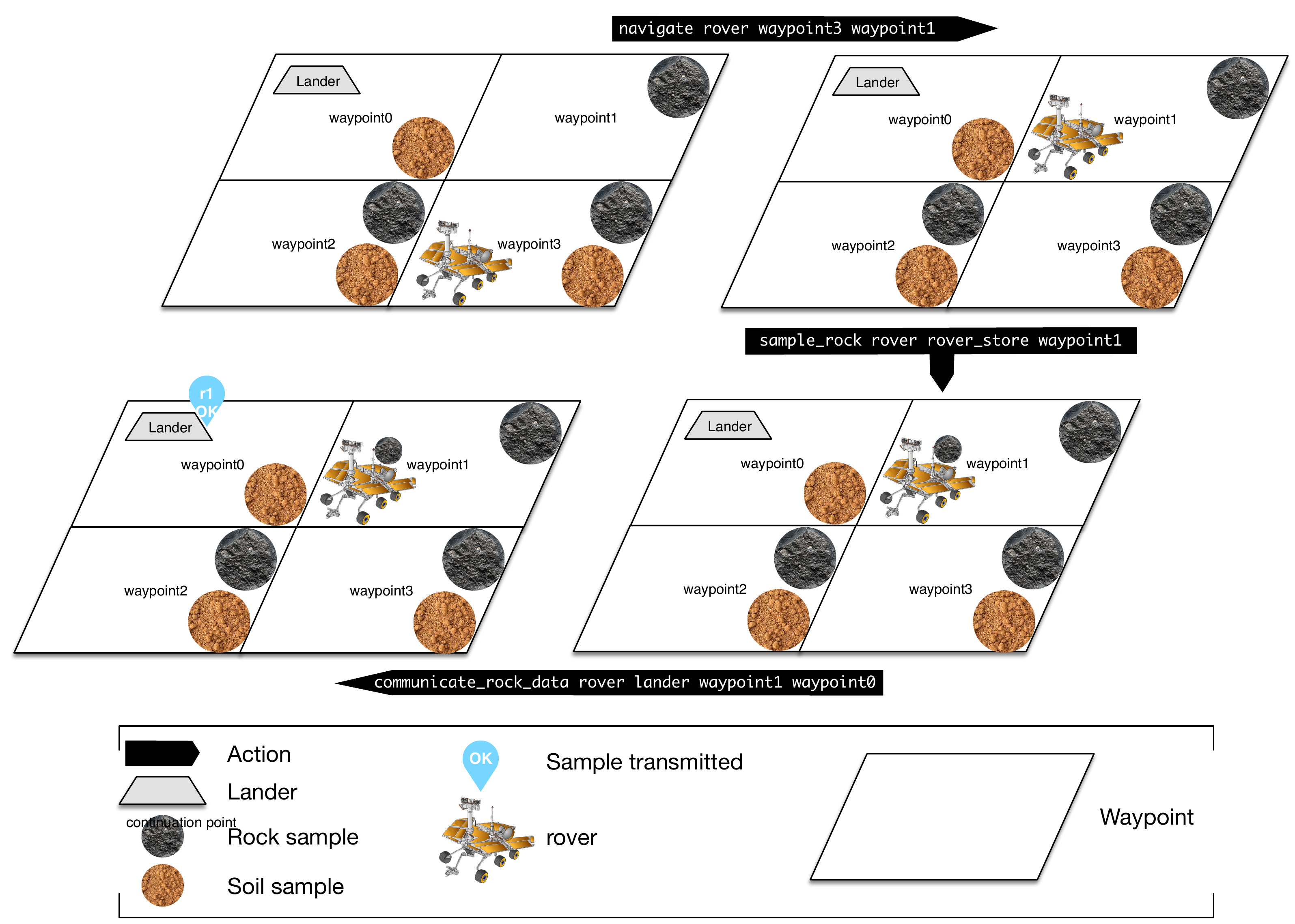}}
\vspace*{8pt}
\caption{Exemple de problème du domaine rover.}
\label{Fig:Example rover domain}
\end{figure}

Nous commençons par définir le problème \textit{rover} que nous avons choisi comme exemple conducteur tout au long de cet article. Cet exemple se représente simplement dans un formalisme HTN et a servi entre autres de benchmark au planificateur SHOP. Ce domaine est une version simplifiée de la mission d'exploration de Mars menée en 2003 par la NASA. Il a aussi été utilisé dans la troisième et la cinquième compétition de planification qui se sont déroulées respectivement en 2002 et 2006. Il traite du déplacement de robots, de l'échantillonnage de sols, de la prise d'image et de la transmission de données de plusieurs rovers. Les robots sont dotés d'un ensemble d'équipements et doivent traverser plusieurs zones appelés \textit{waypoints} afin de collecter des échantillons de sol, de rochers ou prendre une cible en photo et les transmettre à leur base ou {\it lander} à partir d'un waypoint visible depuis la base. La difficulté principale de ce domaine réside dans le fait que chaque rover est prévu pour un certain type de terrain et ne peut donc pas traverser toutes les zones.
La figure~\ref{Fig:Example rover domain} propose une représentation d'un problème très simple extrait du domaine rover où la prise et la transmission d'image sont ignorées. Dans l'état initial, le rover est dans le waypoint3, avec la base située au waypoint0, des échantillons de sol dans les waypoints 0, 2 et 3 ainsi que des échantillons de rochers dans les waypoints 1, 2 et 3. La tâche à réaliser consiste à collecter et transmettre l'échantillon de rocher qui se trouve dans le waypoint1 au lander. Le plan pour résoudre ce problème se présente alors comme suit: Le rover se déplace du waypoint3 vers le waypoint1, collecte l'échantillon de rocher puis le transmet à sa base située au waypoint0.
% Mettre les waypoints en it

%---------------------------------------------------------------
%		Section 3: Le formalisme HTN
%---------------------------------------------------------------

\section{Le formalisme HTN}
\label{le formalisme htn}
L'instanciation des problèmes de planification en général et des problèmes HTN plus particulièrement repose sur des mécanismes agissant sur le problème de planification. Avant d'aborder le processus d'instanciation et sa complexité, nous commençons d'abord par définir le formalisme HTN.

\begin{mydef}
Un problème HTN est un quadruplet $P = (s_0,w,O,M)$ où $s_0$ est l'état initial défini par un ensemble de propositions qui caractérisent le monde, $w$ est un réseau de tâches initial qui définit le but, $O$ est un ensemble d'opérateurs qui définissent les actions qui peuvent être réalisées, et $M$ un ensemble de méthodes qui définissent les décompositions possibles d'une tâche composée en tâches primitives.
\end{mydef}

\begin{mydef}
Un opérateur est un triplet $o=(name(o), pre(o), \textit{eff}(o))$ où $name(o)$ est une expression syntaxique de la forme $t(u_1,...,u_k)$ où $t$ est le nom de l'opérateur et $u_1,...,u_k$ sont des variables ou des constantes typées qui définissent les paramètres de l'opérateur. $pre(o)$ définit les préconditions de l'opérateur qui doivent être vérifiés pour le déclencher. \textit{eff(o)} sont les effets de l'opérateur qui définissent les propriétés générées par l'opérateur. $pre(o)$ et \textit{eff(o)} sont représentés sous la forme d'une expression logique. Leurs sous ensembles {\textit{pre}$^+(o)$, \textit{pre}$^-(o)$, \textit{eff}$^+(o)$, \textit{eff}$^-(op)$} représentent des sous-ensembles positifs et négatifs.
\end{mydef}
Une action $a$ est un opérateur totalement instancié qui définit une fonction de transition permettant de passer d'un état $s$ à un état $s'$ comme suit: $s' = ((s \backslash $\textit{eff}$^-(a)) \cup $\textit{eff}$^+(a))$. $a$ est applicable dans un état $s$ si $pre^+(a) \subseteq s$ et $pre^-(a) \cap s = \emptyset$. Toutes les formules atomiques d'une action sont totalement instanciées et appelées, par abus de langage, \textit{propositions}.
L'exemple \ref{exp op} montre l'opérateur \textit{navigate} du domaine rover, $?x$, $?p1$ et $?p2$ sont ses paramètres.
\begin{exmp}
\label{exp op}
:

\lstset{basicstyle=\footnotesize}

\begin{lstlisting}
(:action navigate
  :parameters (?x - rover ?p1 - waypoint ?p2 - waypoint)
  :precondition
     (and (available ?x) (at ?x ?p1)
          (can_traverse ?x ?p1 ?p2) (visible ?p1 ?p2))
  :effect (and (not(at ?x ?p1)) (at ?x ?p2))
)
\end{lstlisting}
\end{exmp}

\begin{mydef}
Une méthode est un triplet $m=(name(m), subtasks(m), constr(m))$, où $name(o)$ est, comme pour un opérateur, une expression syntaxique de la forme $t(u_1,...,u_k)$ où $t$ est le nom de la méthode et $u_1,...,u_k$ sont des variables ou des constantes typées qui sont utilisées dans la méthode.
\end{mydef}
Une ou plusieurs méthodes peuvent décomposer la même tâche $t(m)$. Chaque méthode représente une façon différente de la réaliser. $subtasks(m)$ est l'ensemble de tâches qui composent $t(m)$ avec un $tag$ qui les remplace dans le reste de la méthode. $constr(m)$ est l'ensemble des contraintes qui portent sur $subtasks(m)$. Le couple $(subtasks(m)$,$constr(m))$ est représenté par un \textit{réseau de tâches}.
\begin{exmp}
\label{exp methode do navigate}
:

\lstset{basicstyle=\footnotesize}
\begin{lstlisting}
(:method do_navigate
  :parameters	(?x - rover ?from ?to - waypoint)
  :expansion
       ((tag t1 (navigate ?x ?from ?mid))
        (tag t2 (visit ?mid))
        (tag t3 (do_navigate ?x ?mid ?to))
        (tag t4 (unvisit ?mid)))
  :constraints
     (and
     	(series t1 t2 t3 t4)
        (before (and (not(can_traverse ?x ?from ?to))
                     (not(visited ?mid))
                     (can_traverse ?x ?from ?mid)) t1)
        (between (visited ?mid) t2 t4)
        (after (and (not(at ?x ?from)) (at ?x ?to)) (t1 t2 t3 t4))
     )
)
\end{lstlisting}
\end{exmp}
L'exemple 3.2 montre une méthode du domaine \textit{rover} nommée \textit{do\_navigate} avec $?x$, $?from$ et $?to$ comme paramètres. Les tâches qui composent \textit{do\_navigate} sont signalées avec le mot clé \textit{:expansion}, et les contraintes portant sur elles par le mot clé \textit{:constraints}.

\begin{mydef}
Une tâche est une expression de la forme $t(u_1,...,u_k)$, où $t$ est le nom de la tâche et $u_1,...,u_k$ l'ensemble de ses paramètres. Lors de la définition du domaine ces paramètres peuvent être soit des variables soit des constantes.

  Une méthode ou un opérateur sont dit \textit{pertinents} pour une tâche $t$ si $t$ est égal à $name(m)$ ou $t$ est égal à $name(o)$. Il existe deux types de tâches: (1) Des tâches composées qui peuvent être décomposées en sous tâches en appliquant une méthode de décomposition, (2) des tâches primitives définies par des opérateurs et qui ne peuvent pas être décomposées.
\end{mydef}

L'action (\textit{navigate rover waypoint3 waypoint2}) de la figure 1 est une version instanciée de la tâche primitive (\textit{navigate ?r ?w1 ?w1}). $navigate$ est le nom de la tâche et $(?r,?w1,?w2)$ ses paramètres. (\textit{rover, waypoint3, waypoint2}) sont les constantes représentant les paramètres de la tâche instanciée. On peut citer comme exemple de tâche composée dans le domaine rover (\textit{get\_soil\_data ?waypoint}) qui se décompose en sous tâches de navigation, de collecte de sol et de transmission de données.

\begin{mydef}
Un réseau de tâches est le tuple $tn = (U,C)$, où $U$ est  un ensemble des tâches et $C$ un ensemble de contraintes qui portent sur ces tâches. Un réseau de tâches ne contenant que des tâches primitives est dit \textit{primitif} et représente un plan solution si toutes les contraintes qu'il contient sont satisfaites.
\end{mydef}
Chaque contrainte représente une condition qui doit être vérifiée dans tous les plans solution. Quatre types de contraintes peuvent être définies dans une méthode:
\begin{itemize}
\item \textit{Order constraint} : une contrainte d'ordre est une expression de la forme \textit{(series $t_1, t_2, ..., t_k$)}. Elle signifie que dans un plan solution, la tâche $t_1$ doit être ordonnée avant la tâche $t_2$, $t_2$ avant $t_3$ ainsi de suite jusqu'à $t_k$.

\item \textit{Before constraint} : une contrainte du type \textit{Before} est une expression de la forme \textit{(before ($\varphi$) ($t_1, ..., t_k$))} où $\varphi$ est une expression logique, et ($t_1, ..., t_k$) est une liste de tâches. Les contraintes \textit{Before} servent à vérifier que l'expression $\varphi$ est vraie dans l'état juste avant la première tâche du groupe ($t_1, ..., t_k$).

\item \textit{After constraint} : les contraintes du type \textit{After} s'écrivent comme des contraintes \textit{Before} sous la forme d'une expression \textit{(after ($\varphi$) ($t_1, ..., t_k$))} et signifient que $\varphi$ doit être vraie dans l'état juste après l'exécution de la dernière tâche de ($t_1, ..., t_k$).
\item \textit{Between constraint} : une contrainte \textit{Between} est une expression de la forme de  \textit{(between ($\varphi$) ($t_{11}, ..., t_{1k}$) ($t_{21}, ..., t_{2k}$))}. L'expression logique $\varphi$ doit être vérifiée dans tous les états entre la dernière tâche de ($t_{11}, ..., t_{1k}$) et la première de ($t_{21}, ..., t_{2k}$).
\end{itemize}

Par rapport au formalisme de la planification classique qui permet de définir les opérateurs, le formalisme HTN rajoute de la connaissance à travers la définition de méthodes de décomposition. On peut voir la définition des méthodes comme une façon de spécifier par des connaissances des heuristiques qui permettent de diriger la recherche. L'utilisation de ces connaissances permet aux algorithmes HTN d'être plus performants.

%---------------------------------------------------------------
%		Section 4: Algorithme d'instanciation du problème HTN
%---------------------------------------------------------------

\section{Algorithme d'instanciation du problème HTN}
L'instanciation des problèmes passe par plusieurs étapes d'énumération et de simplification qui permettent de ne générer que les actions, méthodes et propositions pertinents pour le problème, en réduisant le nombre d'états atteignables et la complexité de la recherche. Pour ce faire, nous nous appuyons sur les travaux de \cite{koehler97IPP} et sur la notion d'inertie introduite par \cite{koehler1999}. Dans cette partie, nous allons commencer par définir le problème de l'instanciation et la complexité qu'implique un tel processus. Ensuite, nous présenterons les deux phases d'instanciation d'un problème HTN, à savoir : (1) la phase d'instanciation des opérateurs. (2) la phase d'instanciation des méthodes.

\subsection{Le problème d'instanciation}
\label{s le probleme instanciation}
La plupart des planificateurs prenant en entrée le langage {\sc PDDL} passent par une étape d'instanciation qui leur permet d'avoir plus d'informations pour effectuer des études d'atteignabilité sur l'espace de recherche afin d'en déduire par exemple des fonctions heuristiques et les utiliser pour améliorer le temps de recherche et la qualité des solutions obtenues.

Le processus d'instanciation consiste à remplacer toutes les occurrences d'une variable typée par les constantes du même type. Le processus d'instanciation génère toutes les instances possibles avec toutes les combinaisons de valeurs constantes possibles. Le problème d'instanciation dépend du nombre de paramètres dans la méthode ou de l'opérateur et de la cardinalité du domaine de chaque variable. Pour un opérateur possédant $k$ paramètres $x_i$ avec $i \in $\{$1,...,k$\} ayant pour domaine $D(x_i)$ et dont la cardinalité est $|D(x_i)|$, la complexité du processus d'instanciation est égale à:
\[\mathcal{O} = \prod_{i=0}^{k} |D(x_i)|\]
On peut voir que le nombre d'instances créées augmente rapidement avec un grand nombre de constantes dans le problème ce qui fait exploser la complexité de l'instanciation et de la recherche. Si on prend comme exemple l'instanciation du domaine \textit{rover} avec le problème \textit{p40} de l'{\sc IPC5}. En considérant juste l'opérateur \textit{communicate\_soil\_data (?x - rover ?l - lander ?p1 - waypoint ?p2 - waypoint ?p3 - waypoint)} on arrive à 14 millions d'instances $= (14 \times rover) \times (1 \times lander) \times (100 \times waypoint1) \times (100 \times waypoint2) \times (100 \times waypoint3)$.

En HTN, le problème de la complexité est d'autant plus important qu'il concerne en plus des opérateurs toutes les méthodes de décomposition. En sachant qu'une méthode contient le plus souvent plusieurs tâches chacune d'entre elles comportant des paramètres, une méthode a donc un nombre de paramètre supérieur ou égal au nombre de paramètres de la tâche qui en contient le plus. Ce qui donne aux méthodes un nombre de paramètre plus important que les opérateurs et donc un nombre d'instances beaucoup plus important. Bien-sur le nombre d'instances de méthodes par rapport au nombre d'opérateurs instanciés dépend de la définition du domaine. Reprenons comme exemple l'instanciation du problème \textit{p40} du domaine \textit{rover}, la méthode \textit{send\_soil\_data (?x - rover ?from - waypoint)} ne compte que deux paramètres déclarés. Cependant, elle contient deux tâches, à savoir: \textit{(do\_navigate ?x ?w1)} et \textit{(communicate\_soil\_data ?x ?l ?from ?w1 ?w2)} qui introduisent trois paramètres supplémentaires. Dans cet exemple, nous arrivons au même nombre d'instances que pour l'opérateur \textit{communicate\_soil\_data}, à savoir 14 millions d'instances. Mais on peut facilement imaginer avoir d'autres tâches et des paramètres supplémentaires.

\subsection{L'instanciation des opérateurs}
\label{s instanciation des op}
L'instanciation des opérateurs passe par quatre étapes : une étape de normalisation des expressions logiques, une étape d’instanciation des opérateurs, une étape de simplification des expressions logiques, et enfin, une étape de simplification et de réduction du nombre d’opérateurs.

\subsubsection{La normalisation des expression logiques}
\label{ss La normalisation des expression logiques}
Dans cette étape, toutes les expressions logiques contenant des implications et des quantificateurs sont reformulées sous une forme conjonctive ou disjonctive simple en suivant les règles de réécriture suivantes :
\begin{itemize}
\setlength\itemsep{0em}
\item $\phi \rightarrow \varphi$ $\Rightarrow$ $\neg\phi \wedge \varphi$
\item $\neg(\phi \wedge \varphi)$ $\Rightarrow$ $\neg\phi \vee \neg\varphi$
\item $forall (?x - type)$ $\Rightarrow$ $x_1 \wedge x_2 \wedge ... \wedge x_n$
\item $exists (?x - type)$ $\Rightarrow$ $x_1 \vee x_2 \vee ... \vee x_n$
\end{itemize}
Toutes les formules logiques contenues dans les opérateurs sont affectées par cette réécriture. Ces règles permettent de mettre les expressions logiques sous la forme conjonctive normale qui est un préalable classique à toute manipulation des expressions logiques.

\subsubsection{L'instanciation des opérateurs}
\label{ss instanciation des op}
Instancier un opérateur consiste à remplacer chaque variable déclarée dans ce dernier par les constantes dont le type correspond à celui de la variable ou en est un sous-type. Pour chaque combinaison de valeurs une nouvelle action est créée. L'exemple \ref{exp action navigate instanciee} montre une instance de l'opérateur \textit{navigate} présente dans l'exemple \ref{exp op} où la variable \textit{?x} à été remplacée par la constante \textit{rover1}, \textit{?p1} par \textit{waypoint3} et \textit{?p2} par \textit{waypoint2}. L'instanciation se base sur les types des variables pour trouver les constantes correspondantes. Mais dans certains cas, les problèmes ne sont pas typés. Il faut alors inférer les types des variables à partir des prédicats unaires dans lesquels elles apparaissent avant d'instancier les opérateurs. Le processus d'inférence des types est présenté en détail dans \cite{koehler1999}.
\begin{exmp}
\label{exp action navigate instanciee}
:

\lstset{basicstyle=\footnotesize}
\begin{lstlisting}
(:action navigate
  :parameters (rover1 waypoint3 waypoint2)
  :precondition
     (and (available rover1) (at rover1 waypoint3)
          (can_traverse rover1 waypoint3 waypoint2)
          (visible waypoint3 waypoint2))
  :effect (and (not(at rover1 waypoint3)) (at rover1 waypoint2))
)
\end{lstlisting}
\end{exmp}

\subsubsection{La simplification des formules atomiques}
\label{ss simplification des formules op}
La phase de simplification doit être effectuée le plus tôt possible dans le processus d'instanciation afin d'optimiser le processus et de réduire sont coût. La simplification consiste à essayer d'évaluer en amont les formules atomiques contenues dans les opérateurs à $vrai$ ou $faux$ en utilisant le concept d'inertie. L'ensemble des propositions considérées comme une inertie regroupe les inerties négatives et les inerties positives.
\begin{itemize}
\item L'inertie négative regroupe les propositions qui n'apparaissent jamais dans les effets négatifs d'un opérateur, donc elles ne sont jamais consommées par une action du problème. Par conséquent, si les propositions considérées comme des inerties négatives apparaissent dans l'état initial, elles seront valides dans tous les états du problème.
\item L'inertie positive regroupe les propositions qui n'apparaissent jamais dans les effets positifs d'un opérateur, donc elles ne sont jamais produites par une action du problème. Donc, si une proposition est une inertie positive et qu'elle n'est pas dans l'état initial, elle n'apparaîtra dans aucun état du problème.
\end{itemize}
Le calcul des inerties est réalisé en parcourant une seule fois l'ensemble des opérateurs du problème. Il se fait très facilement, puisque les effets des opérateurs se limitent à une forme conjonctive. Les prédicats qui ne sont pas dans l'ensemble des inerties sont nommés "\textit{fluent}" et peuvent de ce fait apparaître ou disparaître d'un état à un autre. La simplification des formules atomiques se fait en remplaçant un prédicat dans les inerties positive par $faux$ s'il n'est pas dans l'état initial, et en le remplaçant par $vrai$ s'il est dans les inerties négatives et dans l'état initial. Si $p$ est un prédicat totalement instancié et $I$ l'état initial, alors les règles de simplification sont définies comme suit:
\begin{itemize}
\item \textbf{Si} $p$ est une inertie positive et $p \notin I$ \textbf{alors} $p$ est simplifiée à \textbf{faux}.
\item \textbf{Si} $p$ est une inertie négative et $p \in I$ \textbf{alors} $p$ est simplifiée à \textbf{vrai}.
\item \textbf{Sinon} $p$ ne peut pas être simplifiée.
\end{itemize}
Toutes les propositions simplifiées peuvent être supprimées du problème et toutes celles qui restent sont considérées comme pertinentes. Si on prend, par exemple, le domaine \textit{rover}, la proposition \textit{(can\_traverse ?x - rover ?p1 - waypoint ?p2 - waypoint)} n'apparaît ni dans les effets négatifs ni dans les effets positifs des opérateurs. Elle est par conséquent dans l'ensemble des inerties positives et négatives et peut être remplacée par $vrai$ si elle est dans l'état initial ou par $faux$ sinon.

\subsubsection{La simplification des actions}
\label{ss La simplification des actions}
La simplification des actions vise à trouver celles qui ne pourront jamais être réalisées, en s'appuyant sur les simplifications atomiques explicitées dans §\ref{ss simplification des formules op}. Pour cela, on s'intéresse aux simplifications des expressions logiques définies dans les préconditions et les effets des actions. Comme on l'a vu précédemment, les expressions atomiques peuvent être simplifiées à $vrai$ ou $faux$, ce qui permet de simplifier les préconditions et les effets en appliquant les règles de logique suivantes:
\begin{center}
\framebox[1.1\width][c]{
\begin{tabular}{rclrcl}
   $\neg TRUE$ & $\equiv$ & $FALSE$ & $\neg FALSE$ & $\equiv$ & $TRUE$  \\
   $TRUE \wedge \varphi$ & $\equiv$ & $\varphi$ & $\varphi \wedge \varphi$ & $\equiv$ & $\phi$ \\
   $FALSE \wedge \varphi$ & $\equiv$ & $FALSE$ & $\varphi \vee \varphi$ &$\equiv$& $\varphi$ \\
   $TRUE \vee \varphi$ & $\equiv$ & $TRUE$ & $\varphi \wedge \neg \varphi$ &$\equiv$& $FALSE$ \\
   $FALSE \vee \varphi$ &$\equiv$& $\varphi$ & $\varphi \vee \neg \varphi$ &$\equiv$& $TRUE$ \\

\end{tabular}
}
\end{center}

Si ces simplifications permettent de réduire à \textit{vrai} ou \textit{faux} toute l'expression logique des préconditions ou des effets d'une action, elle peut être simplifiée comme suit:
\begin{itemize}
\setlength\itemsep{0em}
\item Si la précondition ou un effet d'une action est remplacé par $faux$, l'action est supprimée du problème de planification. Dans le cas où la précondition est fausse, l'action ne peut jamais être appliquée. Dans le cas où c'est l'effet est simplifié à \textit{faux}, l'action produit un état incohérent.
\item Si tous les effets d'une action sont évalués à $vrai$, elle peut être supprimée parce qu'elle ne produit aucun changement.
\end{itemize}

\subsection{L'instanciation des méthodes}
\label{ss instanciation des meth}
L'instanciation des méthodes passe par cinq étapes : Une étape de normalisation des expressions contenues dans les contraintes, une étape d’inférence de types, une étape d’instanciation des méthodes, une étape de simplifications des expressions logiques, et enfin, une étape de simplification et de réduction des méthodes instanciées.

\subsubsection{La normalisation des expressions logiques}
\label{ss La normalisation des expressions logiques}
En HTN, la normalisation des expressions logiques utilise les règles définies dans §\ref{ss La normalisation des expression logiques} pour réécrire les expressions logiques contenues dans les contraintes des méthodes. Les contraintes concernées par cette normalisation sont celles contenant des expressions logiques: \textit{Before}, \textit{After} et \textit{Between}. La réécriture des expressions logiques sous forme normale conjonctive est nécessaire pour les futures manipulations.

\subsubsection{L'inférence des types}
\label{ss inference des types}
La différence avec l'instanciation des opérateurs réside dans le fait que les méthodes permettent de spécifier des variables non déclarées dans les paramètres. Dans l'exemple 3.2, la variable \textit{?mid} est utilisée dans les sous tâches et presque toutes les contraintes mais n'est pas déclarée dans les paramètres de la méthode. Ces variables n'ont aucun type défini, ce qui nécessite d'effectuer l'inférence des types de ces variables avant de pouvoir réaliser l'instanciation des méthodes. Le processus d'inférence pour chaque variable non déclarée se découpe en deux étapes :

\paragraph{L'inférence à partir des tâches}
\begin{enumerate}
\setlength\itemsep{0em}
\item Récupérer toutes les tâches $T$ qui contiennent dans leurs paramètres la variable non déclarée,
\item Pour chaque tâche $t \in T$, récupérer l'opérateur $op_r$ ou les méthodes $m_r$ qui sont pertinents pour $t$.
\item Pour chaque tâche $t \in T$, récupérer les types déclarés dans les paramètres de $op_r$ ou des $m_r$. Si on obtient deux types $A$ et $B$, où $B$ est un sous type de $A$, on ne garde que le type $B$,
\item Si plusieurs types n'ayant aucun lien d'héritage sont récupérés, alors une erreur de typage qui doit être signalée.
\end{enumerate}

\paragraph{L'inférence à partir des contraintes}
\begin{enumerate}
\setlength\itemsep{0em}
\item Récupérer l'ensemble des propositions $P$, à partir des contraintes qui contiennent dans leurs paramètres la variables non déclarée,
\item Pour chaque proposition $p \in P$, récupérer les types de la variable non déclarée à partir de la formule atomique pertinente pour $p$,
\item Si plusieurs types qui n'ont aucun lien d'héritage sont récupérés, alors une erreur de typage doit être signalée. Sinon, si les types ont un lien d'héritage, il faut garder le type qui n'a pas de sous types.
\end{enumerate}
À la fin, si, à partir des différentes étapes d'inférence, plusieurs types sont obtenus, il faut garder le type ne possédant pas de sous types. S'il reste encore plusieurs types, alors c'est une erreur.

\subsubsection{L'instanciation des méthodes}
\label{ss instanciation des methodes}
L'instanciation d'une méthode consiste à remplacer les variables qu'elle contient par toutes leurs instances possibles. Comme pour les opérateurs, une instance d'une variable est une constante dont le type est égal à celui de la variable ou en est un sous-type. Chaque instance de méthode correspond à une combinaison d'instances des variables qu'elle contient.
\begin{exmp}
\label{exp methode do navigate instanciee}
:

\lstset{basicstyle=\footnotesize}
\begin{lstlisting}
(:method do_navigate
  :parameters	(rover1 waypoint3 waypoint0)
  :expansion
       ((tag t1 (navigate rover1 waypoint3 waypoint1))
        (tag t2 (visit waypoint1))
        (tag t3 (do_navigate rover1 waypoint1 waypoint0))
        (tag t4 (unvisit waypoint1)))
  :constraints
     (and
     	(series t1 t2 t3 t4)
        (before (and (not(can_traverse rover1 waypoint3 waypoint0))
                     (not(visited waypoint1))
                     (can_traverse rover1 waypoint3 waypoint1)) t1)
        (between (visited waypoint1) t2 t4)
        (after (and (not(at rover1 waypoint3))
                    (at rover1 waypoint0)) (t1 t2 t3 t4))
     )
)
\end{lstlisting}
\end{exmp}
Le résultat de l'instanciation de la méthode \textit{do\_navigate} de l'exemple \ref{exp methode do navigate} est la méthode totalement instanciée présentée dans l'exemple \ref{exp methode do navigate instanciee} avec la combinaison suivante: \textit{?x $ \equiv$ rover1}, \textit{?from $ \equiv$ waypoint3}, \textit{?to $ \equiv$ waypoint0}, \textit{?mid $ \equiv$ waypoint1}. La valeur affectée à \textit{?mid} correspond au type \textit{waypoint} en s'appuyant sur l'inférence des types depuis les sous tâches et les contraintes de la méthode.

\subsubsection{La simplification des formules atomiques}
\label{ss simplification des formules atomiques des methode}
En sachant que le nombre d'instances de méthodes obtenues pour un problème est plus grand que celui des opérateurs, la phase de simplification doit être effectuée le plus tôt possible dans le processus d'instanciation. Comme pour les opérateurs, la simplification consiste à essayer d'évaluer en amont les formules atomiques contenues dans les contraintes de la méthode à $vrai$ ou $faux$ en utilisant le principe d'inertie. Sans redéfinir la notion l'inertie et les règles de simplification présentés dans §\ref{ss simplification des formules op}, le résultat conduirait à la simplification de la contraintes définies dans la méthode \textit{do\_navigate} de l'exemple \ref{exp methode do navigate instanciee}:
\lstset{basicstyle=\footnotesize}
\begin{lstlisting}
(before (and (not(can_traverse rover1 waypoint3 waypoint0))
             (not(visited waypoint1))
             (can_traverse rover1 waypoint3 waypoint1)) t1)
)
\end{lstlisting}
par:
\lstset{basicstyle=\footnotesize}
\begin{lstlisting}
(before (false) t1))
\end{lstlisting}
si \textit{(can\_traverse rover1 waypoint3 waypoint0)} est dans l'état initial. On voit que le processus de simplification des formules atomiques réduit considérablement la complexité des contraintes et réduit de ce fait la complexité du traitement effectué lors de la phase de simplification des méthodes.

\subsubsection{La simplification des méthodes}
\label{ss simplification des methodes}
La simplification des méthodes cherche à identifier et supprimer les méthodes dont les contraintes ne vont jamais être vérifiées quel que soit l'état initial et le but du problème. Deux simplifications sont réalisées lors d'un seul parcours sur l'ensemble des méthodes :

\paragraph{La simplification fondée sur les contraintes}
Comme son nom l'indique, la simplification fondée sur les contraintes s'appuie sur les évaluations de leurs expressions logiques. De la même manière qu'elles peuvent l'être pour les opérateurs, ces dernières peuvent être simplifiées à $vrai$ ou $faux$ en se fondant sur les règles de logiques présentées dans §\ref{ss La simplification des actions}. En sachant que les contraintes définies dans les méthodes sont sous forme normale conjonctive, les règles de simplifications qui s'appliquent sont les suivantes :
\begin{itemize}
\setlength\itemsep{0em}
\item Si l'expression logique est simplifiée à $vrai$, toute la contrainte est supprimée, puisqu'elle est toujours vérifiée.
\item Si la formule logique d'une contrainte est simplifiée à $faux$, toute la méthode est supprimée. Dans ce cas, cette contrainte ne pourra jamais être vérifiée dans le problème, ce qui implique que cette méthode ne mènera jamais à un plan solution.
\end{itemize}

\paragraph{La simplification fondée sur les tâches}
La simplification fondée sur les tâches cherche à supprimer les méthodes dont les tâches primitives ne peuvent pas être réalisées. En sachant que la simplification des opérateurs est effectuée avant celle des méthodes, la procédure de simplification est comme suit :
\begin{enumerate}
\setlength\itemsep{0em}
\item Récupérer l'ensemble des tâches primitives $T$ définies dans la méthode à simplifier,
\item Pour chaque tâche $t \in T$, vérifier si l'action permettant de réaliser $t$ a été supprimée lors de la phase de simplification des opérateurs. Si c'est le cas, alors toute la méthode est supprimée. Sinon, elles est conservée,
\item Si aucune action n'est trouvée, toute la méthode est supprimée.
\end{enumerate}
Une simplification des méthodes fondée sur les tâches pourrait être réalisée en considérant les tâches composées. Dans cette simplification, une méthode est supprimée si elle contient une tâche composée dont l'ensemble des méthodes pertinentes est vide. Ce processus de simplification nécessiterait plusieurs itérations sur l'ensemble des méthodes. Dans chaque itération, des méthodes sont supprimées parce qu'elles contiennent des tâches composées dont l'ensemble des méthodes pertinentes est vide. En sachant que ces méthodes peuvent être pertinentes pour des tâches $t'$, leur suppression ouvre la possibilité de supprimer d'autres méthodes contenant $t'$ dans la prochaine itération. Le processus se poursuit ainsi jusqu'à ce que l'ensemble des méthodes à simplifier se stabilise.

\section{Tests et résultats}
Après avoir présenté la méthode d'instanciation et de simplification des domaines HTN, nous cherchons à savoir dans cette partie si la diminution du nombre des méthodes et des opérateurs, en utilisant une approche instanciée, améliore les performances des algorithmes HTN. Pour répondre à cette question, nous avons implémenté une version simplifiée de l'algorithme SHOP nommée \textit{iSHOP (instanciated SHOP)} et fonctionnant avec un problème totalement instancié, et nous l'avons comparé avec l'algorithme SHOP classique.

\subsection{L'algorithme iSHOP}
iSHOP est développé en Java en utilisant la librairie de planification PDDL4J \cite{pddl4j}. La librairie inclut les modules nécessaires pour l’analyse lexicale et syntaxique et la planification classique en plus du module d’instanciation et de simplification des problèmes de planification HTN que nous avons développé. Notre choix s'est porté sur SHOP parce que c'est l'algorithme de référence pour la planification HTN, qu'il est performant et simple à implémenter.

L'algorithme 1 présente un processus générique de iSHOP qui prend en entrée le problème \textit{(S,T,$O$,$M$,$P$)} où $S$ est un état, \textit{T = $(t_1,t_2,...,t_n)$} est une liste de tâches, $O$ est l'ensemble des actions, $M$ les méthodes instanciées et $P$ les propositions du problème. Dans cet algorithme on commence par vérifier si l'ensemble des tâches $T$ n'est pas vide, sinon il faut retourner \textit{un plan vide}. On récupère ensuite la première tâche et selon si elle est primitive ou composée, on applique l'action correspondante à l'état dans le premier cas, ou on applique une méthode de décomposition dans le deuxième. Le processus se répète jusqu'à ce que toutes les tâches du réseau de tâches sont traitées. Le plan est constitué des actions appliquées aux tâches primitives du réseau de tâches. Contrairement à SHOP, aucune contrainte d'instanciation n'est maintenue et la vérification des préconditions est effectuée en une seule vérification d'inclusion dans l'état, grâce à l'instanciation qui nous permet de représenter les préconditions sous forme d'un ensemble de propositions à comparer avec l'ensemble des propositions de l'état.
 \begin{algorithm}[H]
 \label{algorithme iSHOP}
    \caption{iSHOP($S,T,O,M,P$)}\label{euclid}
    \begin{algorithmic}[1]
    \If {$T$ = $\emptyset$} \Return plan vide
    \EndIf
      \State $t \gets$ la première tâche de $T$
      \State $U \gets T - t$
      \If{$t$ est primitive}
      	\State $a \gets$ l'action pertinente pour $t$
        \If{$precond(a) \in S$}
        	\State $P \gets$ iSHOP($a(S),U,O,M,P$)
            \State \Return $P$
        \Else
        	\State \Return Échec
        \EndIf
      \Else
        \State $active \gets$ \{$m \in M$ et $m$ est pertinente pour $t$\}
        \If{$active \neq \emptyset$}
          \State Choisir de façon non déterministe $m \in active$
          \If{$precond(m) \in S$}
          	\State $T' \gets U \cup tasks(m)$
              \State iSHOP($S,T',O,M,P$)
          \Else
          	\State \Return Échec
          \EndIf
        \Else
          \State \Return Échec
        \EndIf
       \EndIf
    \end{algorithmic}
  \end{algorithm}
iSHOP utilise la structure des méthodes définie dans §\ref{le formalisme htn}. Cette structure permet de définir les méthodes utilisées dans l'algorithme SHOP mais reste beaucoup trop expressive. Pour iSHOP toutes les contraintes sont ignorées sauf les \textit{before} qui représentent les préconditions définies dans SHOP. L'ordre des tâches suit l'ordre dans lequel elles ont été définies dans les méthodes et le problème. L'expressivité du langage d'entrée permet une évolution facile à SHOP2, en considérant les contraintes d'ordre, ou vers les autres algorithmes HTN avec toutes les contraintes. Dans les problèmes, en plus de définir les \textit{goal task}, iSHOP permet en plus de définir un état but avec les contraintes \textit{after}. Chaque nœud de l'arbre de recherche contient un task network et l'état du système. Il représente l'état atteint après l'exécution de la dernière action exécutable dans le task network. Elle correspond à la dernière tâche primitive de la suite de tâches primitive sans interruption en partant du début.

\subsection{Le cadre expérimental}
Dans le cadre de notre comparaison entre SHOP et iSHOP, nous avons effectué une première compagne de tests où nous avons comparé les deux algorithmes sur trois domaines de planification: rover, childsnack et satellite. La première phase de test s'est limitée aux trois problèmes cités précédemment, principalement à cause de la difficulté de définir des méthodes HTN dans deux langages différents, un pour SHOP et un pour iSHOP, tout en veillant à ce que les deux définitions soient les plus proches possibles afin de ne pas fausser la comparaison. En plus des décompositions, nous avons réécrit en HTN tous les problèmes donnés dans les compétitions de planification pour trois domaines, ce qui nous a permis de comparer aussi les deux algorithmes avec un algorithme de planification classique qui a déjà été utilisé dans les compétitions de planification, à savoir \textit{Fast Downward} \cite{helmert2006fast}. Afin de recréer exactement les mêmes conditions pour les deux algorithmes, nous avons voulu avoir les deux algorithmes codés dans le même langage. Donc, nous avons choisi de comparer iSHOP qui est codé en Java avec un planificateur développé au sein de l'équipe et implémentant SHOP en Java. Nous n'avons pas utilisé JSHOP2, puisqu'il implémente l'algorithme SHOP2 et montre en plus des erreurs d'exécution sur le domaine rover.
\paragraph{Matériel}
Les tests ont été effectués sur un ordinateur multi-cœurs Intel Core i7 cadencés à 2,2GHZ possédant 16 Go de RAM DDR3 avec 1600MHz. Le nombre de cœurs n'affecte pas beaucoup les performances des algorithmes, puisqu'ils n'implémentent pas des techniques de parallélisation et que le processus Java s'exécute sur un seul cœur. Les autres cœurs ne sont utilisés par la JVM que pour la compression de mémoire lorsque celle qui est allouée est dépassée. La mémoire allouée pour l'exécution du processus Java dans cet expérimentation est de 10 Go.
\paragraph{Critères de comparaison}
Le but de l'instanciation est de réduire le temps de recherche. Nous avons, donc, comparé les algorithmes en nous appuyant sur les critères de la catégorie \textit{agile} des compétitions de planification, où chaque planificateur obtient un score basé sur son temps de traitement par rapport au meilleur temps des autres algorithmes. Pour iSHOP, le temps total est composé du temps de recherche et du temps de pré-traitement. Nous nous sommes aussi intéressés à la longueur des plans obtenus avec chaque algorithme. Mais la comparaison en terme de longueur de plan reste très liée à la définition du domaine, surtout entre les algorithmes HTN et classiques.

La figure \ref{fig:fig1} présente les résultats obtenus lors de la première compagne de tests en terme de temps d'exécution. Sur l'axe des abscisses, sont représentés les problèmes de planification des trois domaines. L'axe des ordonnées représente le temps, en secondes, pris par chaque planificateur pour trouver un résultat. Si aucun point n'est affiché pour un problème donné, cela signifie que le planificateur n'a pas trouvé de solution dans le temps imparti qui est de 10 minutes. La figure 3 représente les résultats des algorithmes en terme de longueur de plans qui sont représentés en nombre d'actions sur l'axe des ordonnées.

\subsection{Résultats et évaluation}
Les résultats obtenus avec un algorithme planification HTN restent très liés à la définition du domaine. En prenant en compte cette caractéristique et dans un esprit d'équité, nous avons défini des domaines HTN très similaires pour les deux algorithmes. Néanmoins, de légères différences n'ont pas pu être évitées, cela est dû principalement aux spécificités des langages d'entrée de chaque planificateur.

\begin{center}
\begin{figure}[t]
\begin{subfigure}{.5\textwidth}
  \centering
  \includegraphics[width=1\linewidth]{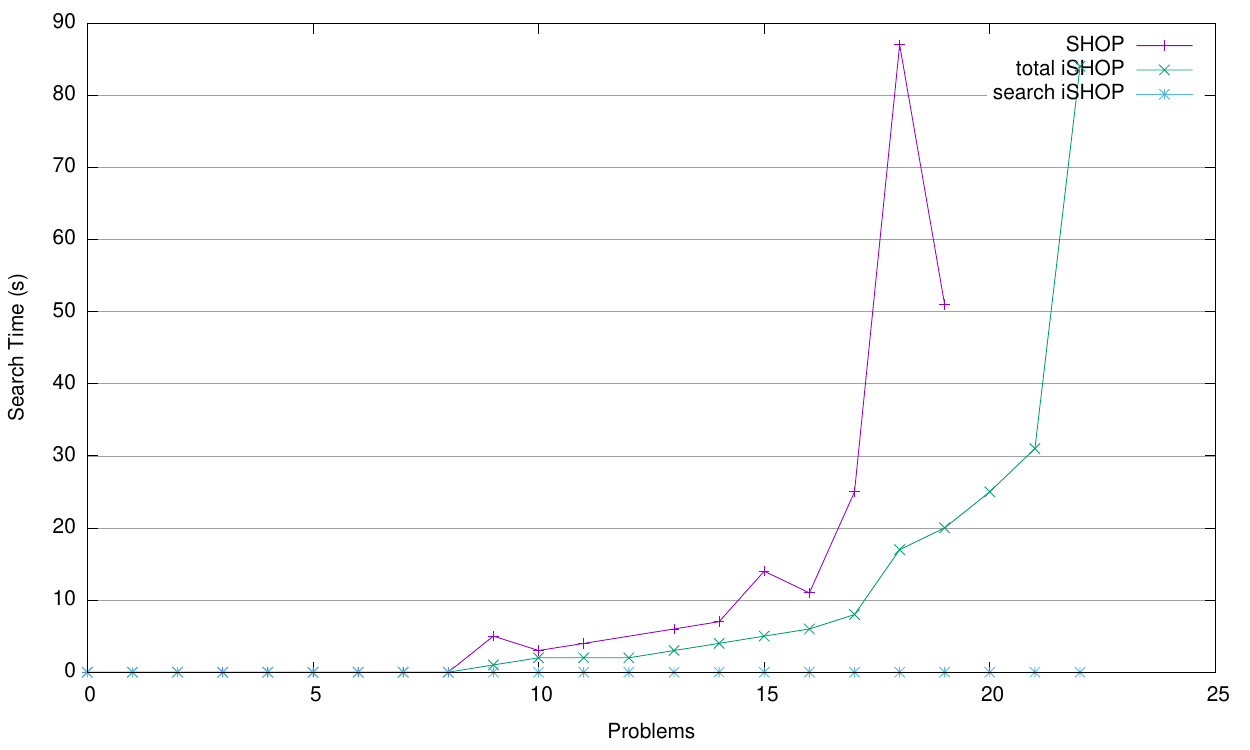}
  \caption{\scriptsize{Temps de traitement de SHOP et iSHOP sur le domaine rover}}
  \label{fig:sfig11}
\end{subfigure}
\begin{subfigure}{.5\textwidth}
  \centering
  \includegraphics[width=1\linewidth]{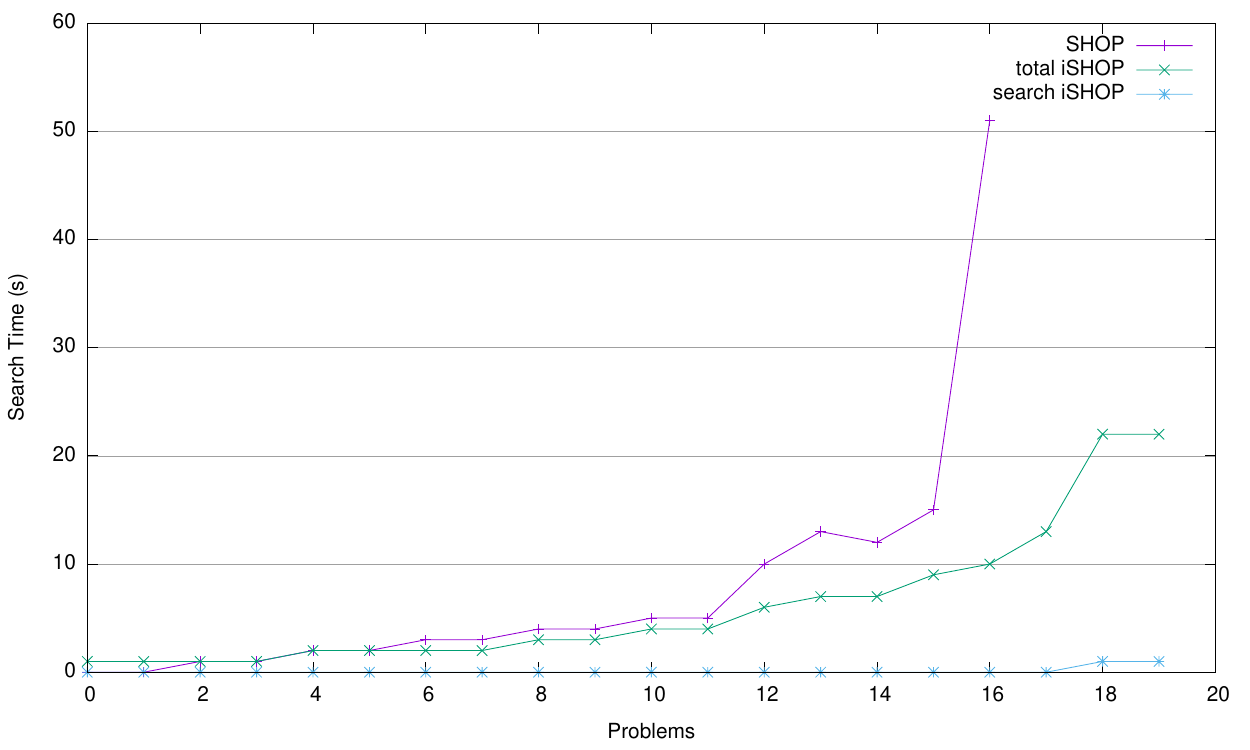}
  \caption{\scriptsize{Temps de traitement de SHOP et iSHOP sur le domaine childsnack}}
  \label{fig:sfig12}
\end{subfigure}
\begin{subfigure}{.5\textwidth}
  \centering
  \includegraphics[width=1\linewidth]{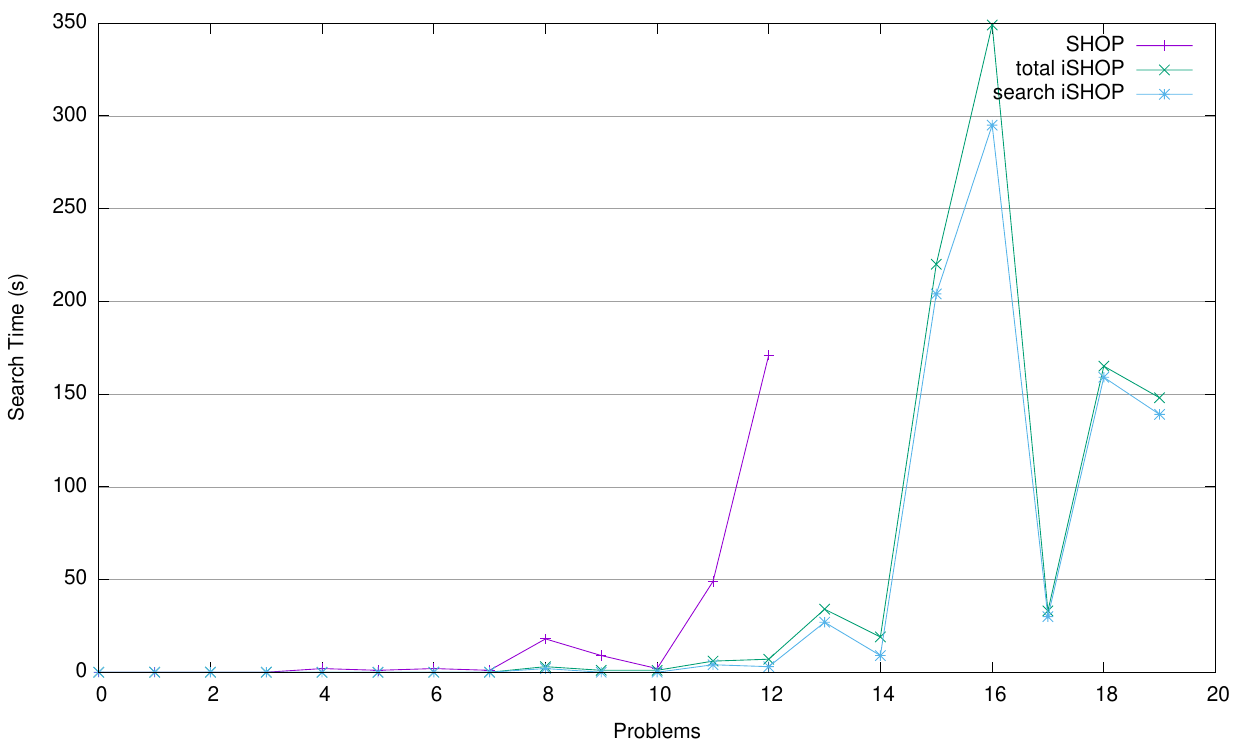}
  \caption{\scriptsize{Temps de traitement de SHOP et iSHOP sur le domaine satellite}}
  \label{fig:sfig13}
\end{subfigure}
\begin{subfigure}{.5\textwidth}
  \centering
  \includegraphics[width=1\linewidth]{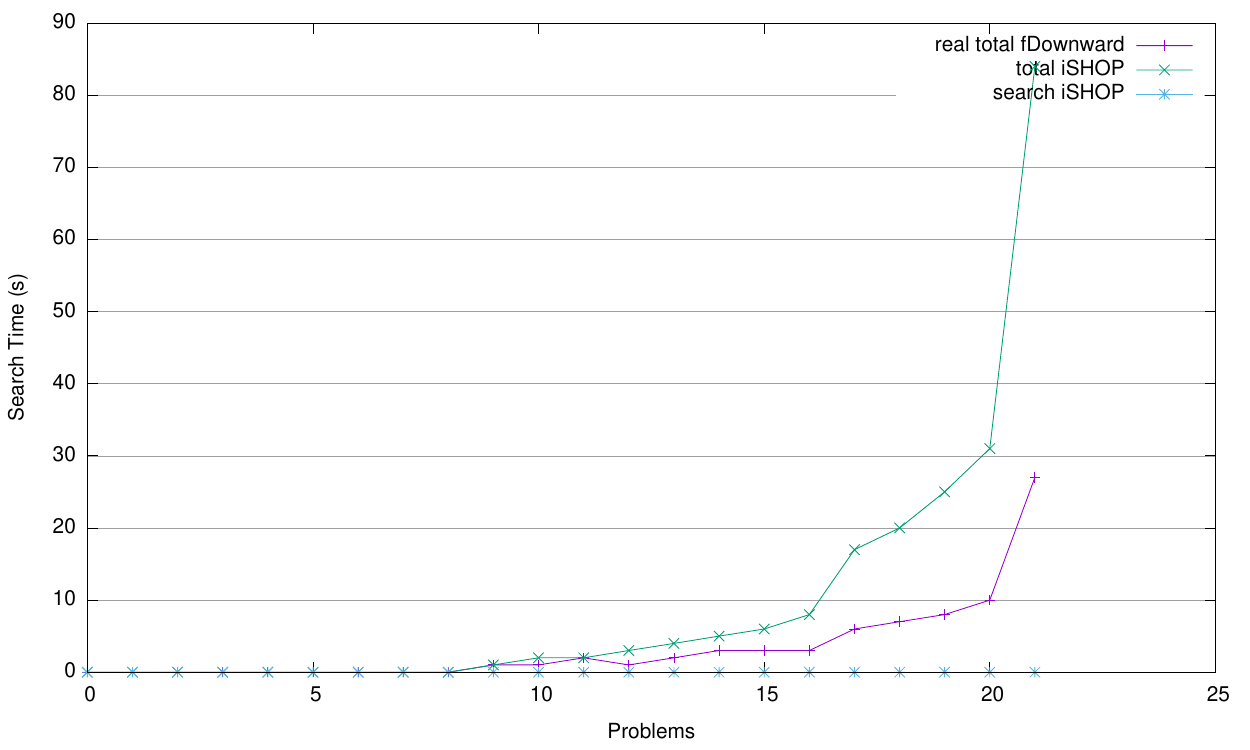}
  \caption{\scriptsize{Temps de traitement de f\_downward et iSHOP sur le domaine rover}}
  \label{fig:sfig14}
\end{subfigure}
\caption{Comparaison des temps de traitement des algorithmes iSHOP, Fast Downward et SHOP sur les domaines de planification: rover, childsnack et satellite}
\label{fig:fig1}
\end{figure}
\end{center}

En considérant les temps de recherche affichés dans la figure \ref{fig:fig1} (a, b et c), on peut remarquer que, iSHOP prend moins de temps que SHOP pour trouver une solution. L'écart de temps de recherche augmente avec la complexité des problèmes. Il est très petit, voire négligeable sur les problèmes simples, de l'ordre de plusieurs secondes, sur les problèmes moyens et de l'ordre de la dizaines de secondes sur les problèmes complexes. On peut voir également que SHOP ne trouve plus de solutions à partir du 20e problème dans rover, du 17e dans childsnack et du 12e dans satellite, alors que iSHOP trouve une solution pour le problème le plus complexe de rover en 85 sec, de childsnack en 22 sec et satellite en 350 sec.

On peut noter que le temps du pré-traitement représente plus de 90\% du temps total de traitement de iSHOP dans childsnack et satellite, et moins de 10\% du temps total dans satellite. Cela est dû, en grande partie, à la définition des méthodes et aux inerties présentes dans le problème. Dans rover, nous avons suivi la définition du domaine donnée par SHOP mais nous aurions pu définir le domaine autrement, ce qui aurait conduit à plus de simplifications. Dans le graphe (d) de la figure 2, la mauvaise performance de iSHOP par rapport à Fast Downward est due principalement au temps de pré-traitement pour les mêmes raisons citées précédemment. Sans le temps de prétraitement, le temps de recherche de iSHOP est de 0,78 sec avec moins de 7 844 nœuds explorés pour le problème le plus complexe, contre 8,21 sec et 7 091 nœuds explorés avec Fast Downward.

Le tableau 1 montre les scores obtenus pour chaque algorithme sur les trois domaines en s'appuyant sur les règles de la catégorie \textit{agile} de l'IPC 8. On voit que sur le domaine rover, Fast Downward est en tête avec un score parfait de 22/22. Il est 13,95\% plus performant que iSHOP, et SHOP est 16,4\% plus performant que SHOP. Sur le domaine childsnack, iSHOP est en tête avec un score 18,59/20. Il est 62,5\% plus performant que Fast Downward. Sur le domaine satellite, c'est Fast Downward qui a réalisé le meilleure score avec 18,65/20. iSHOP est 6,23\% moins performant que Fast Downward et 43\% plus performant que SHOP. La note globale sur les trois problèmes pour Fast Downward, iSHOP et SHOP est respectivement de: 45,5/62 54,81/62 et 37,57/62. Sur les trois domaines, iSHOP est 15\% plus performant que Fast Downward et 27,8\% plus performant que SHOP. Les résultats prouvent que iSHOP reste plus performant en terme de temps de calcul que la version SHOP classique avec une nette avance sur les problèmes complexes.

Comme pour le temps de recherche, la longueur des plans, en HTN dépend grandement de la définition des méthodes de décomposition. Dans la figure 3 (a,b,c), on peut observer que la longueur des plans obtenus avec SHOP et iSHOP. On voit clairement que les deux algorithmes trouvent des plans de longueurs assez similaires. Dans le cas de Fast Downward, la différence de plans avec iSHOP est très grande. Cela est dû principalement à l'approche HTN de iSHOP qui définit le but comme une succession de tâches ordonnées, alors que Fast Downward, en plus d'heuristiques très performantes, définit le but comme un état à atteindre, où l'ordre des actions n'est pas restreint par la définition du problème.

\begin{table}
\centering
\begin{tabular}{|p{0.7cm}lll|}
\hline
Rover		& fdwd		& iSHOP		& SHOP	\\
\hline
pb00 		&	1 		&	1		& 1		\\
pb01 		&	1 		&	1		& 1		\\
pb02 		&	1 		&	1		& 1		\\
pb03 		&	1 		&	1		& 1		\\
pb04 		&	1 		&	1		& 1		\\
pb05 		&	1 		&	1		& 1		\\
pb06 		&	1 		&	1		& 1		\\
pb07 		&	1 		&	1		& 1		\\
pb08 		&	1 		&	1		& 1		\\
pb09 		&	1		&	0,88	& 0,63	\\
pb10 		&	1		&	0,86	& 0,79	\\
pb11 		&	1		&	0,96	& 0,81	\\
pb12 		&	1		&	0,72	& 0,63	\\
pb13 		&	1		&	0,77	& 0,64	\\
pb14 		&	1		&	0,81	& 0,59	\\
pb15 		&	1		&	0,74	& 0,62	\\
pb16 		&	1		&	0,73	& 0,54	\\
pb17 		&	1		&	0,70	& 0,47	\\
pb18 		&	1		&	0,70	& 0,54	\\
pb19 		&	1		&	0,67	& 0		\\
pb20 		&	1		&	0,67	& 0		\\
pb21 		&	1		&	0,67	& 0		\\
\hline
Total 		&	22		&	18,93	& 15,32	\\
\hline
\end{tabular}
\caption{Scores des algorithmes Fast Downward, iSHOP et SHOP sur le domaine Rover}
\end{table}

\begin{table}
\centering
\begin{tabular}{|p{0.7cm}lll|}
\hline
Childs&		fdwd	& iSHOP		& SHOP	\\
\hline
pb00	&		0		& 0,88		& 1		\\
pb01	&		0,49	& 0,91		& 1		\\
pb02	&		0,34	& 0,94		& 1		\\
pb03	&		1		& 0,61		& 0,61	\\
pb04	&		1		& 0,64		& 0,66	\\
pb05	&		1		& 0,70		& 0,72	\\
pb06	&		0		& 1		 	& 0,95	\\
pb07	&		0		& 1		 	& 0,95	\\
pb08	&		0		& 1		  	& 0,91	\\
pb09	&		1		& 0,88		& 0,84	\\
pb10	&		0		& 1		 	& 0,92	\\
pb11	&		0		& 1		  	& 0,93	\\
pb12	&		0		& 1		 	& 0,83	\\
pb13	&		0		& 1		 	& 0,81	\\
pb14	&		0		& 1		 	& 0,83	\\
pb15	&		0		& 1		  	& 0,81	\\
pb16	&		0		& 1		 	& 0,59	\\
pb17	&		0		& 1		 	& 0		\\
pb18	&		0		& 1		 	& 0		\\
pb19	&		0		& 1		 	& 0		\\
\hline
Total	&		4,84	& 18,59		& 14,42	\\
\hline
\end{tabular}
\caption{Scores des algorithmes Fast Downward, iSHOP et SHOP sur le domaine Childsnack}
\end{table}

\begin{table}
\centering
\begin{tabular}{|p{0.7cm}lll|}
\hline
Satlte&		fdwd		& iSHOP		& SHOP	\\
\hline
pb00	&		1 			& 1    	& 1			\\
pb01	&		1 			& 1    	& 1			\\
pb02	&		1 			& 1    	& 1			\\
pb03	&		1 			& 1    	& 1			\\
pb04	&		1 			& 1    	& 0,33		\\
pb05	&		1 f			& 1    	& 0,39		\\
pb06	&		1 			& 1    	& 0,37		\\
pb07	&		1 			& 1    	& 0,37		\\
pb08	&		1 			& 0,43  & 0,32		\\
pb09	&		1 			& 0,66  & 0,42		\\
pb10	&		0,86		& 1    	& 0,76		\\
pb11	&		1			& 0,61  & 0,39		\\
pb12	&		0,91		& 1    	& 0,42		\\
pb13	&		0,98		& 1    	& 0			\\
pb14	&		0,88		& 1    	& 0			\\
pb15	&		1			& 0,90  & 0			\\
pb16	&		0			& 1    	& 0			\\
pb17	&		1			& 0,48  & 0			\\
pb18	&		1			& 0,69  & 0			\\
pb19	&		1			& 0,47  & 0			\\
\hline
Total	&		18,65	& 17,28	& 7,81		\\
\hline
\end{tabular}
\caption{Scores des algorithmes Fast Downward, iSHOP et SHOP sur sur le domaine Satellite}
\end{table}

\begin{center}
\begin{figure}[h]
\begin{subfigure}{.5\textwidth}
  \centering
  \includegraphics[width=1\linewidth]{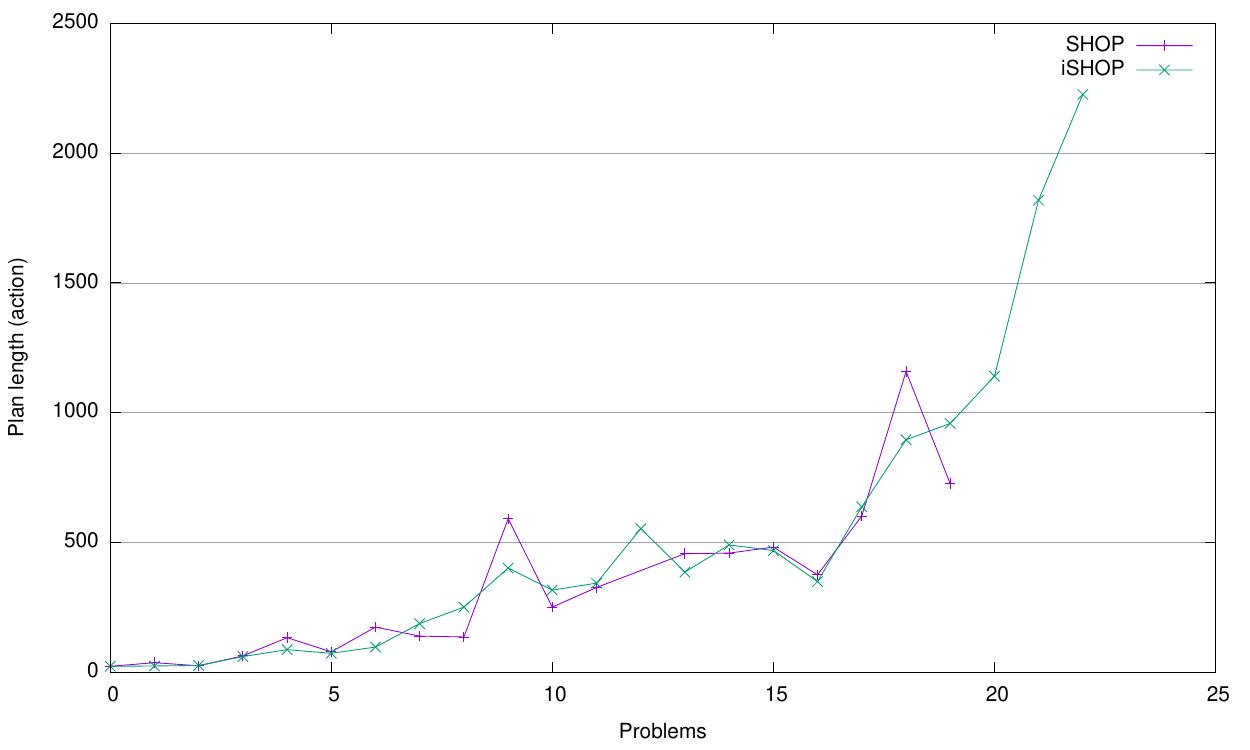}
  \caption{\scriptsize{Longueur des plans de SHOP et iSHOP sur le domaine rover}}
  \label{fig:sfig21}
\end{subfigure}
\begin{subfigure}{.5\textwidth}
  \centering
  \includegraphics[width=1\linewidth]{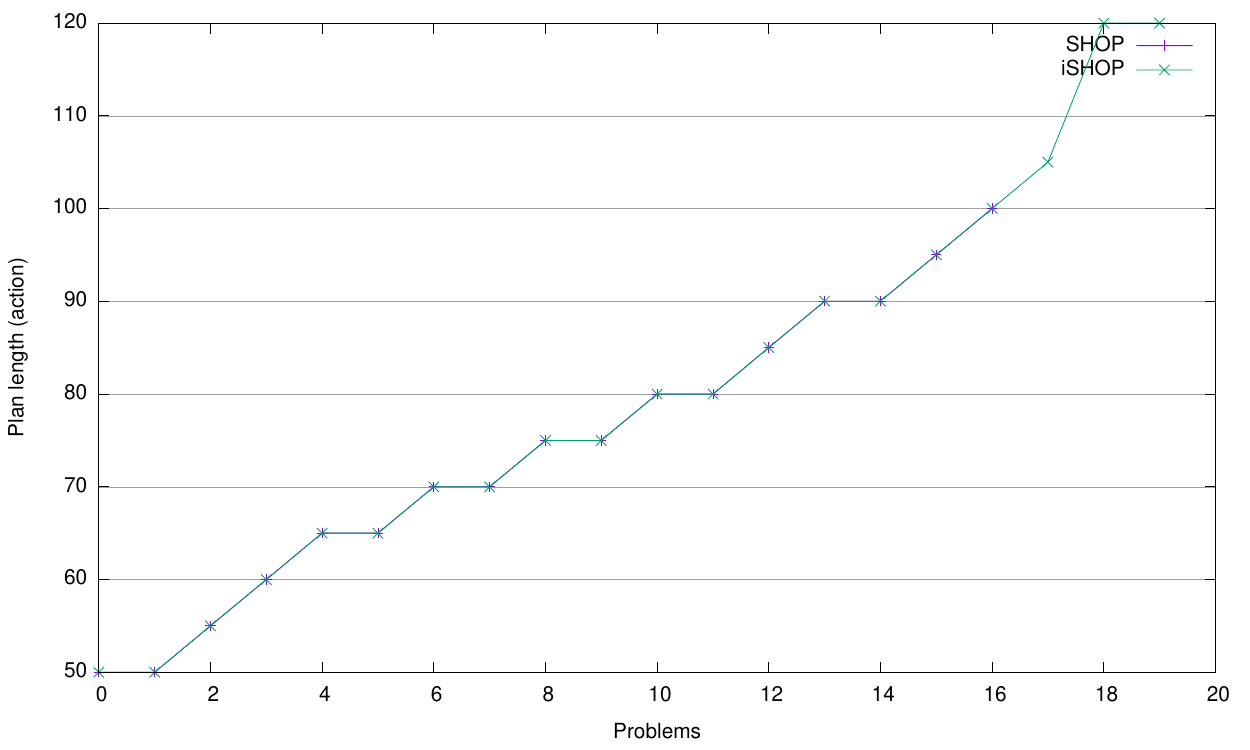}
  \caption{\scriptsize{Longueur des plans de SHOP et iSHOP sur le domaine childsnack}}
  \label{fig:sfig22}
\end{subfigure}
\begin{subfigure}{.5\textwidth}
  \centering
  \includegraphics[width=1\linewidth]{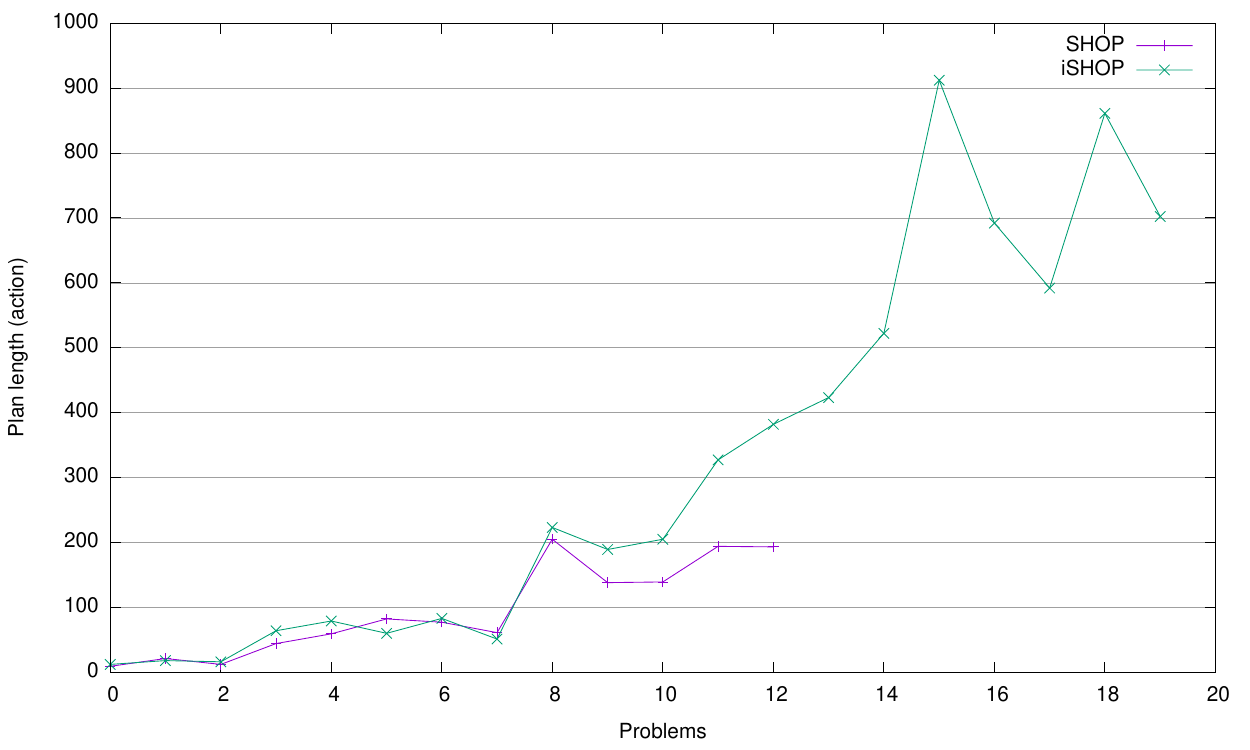}
  \caption{\scriptsize{Longueur des plans de SHOP et iSHOP sur le domaine satellite}}
  \label{fig:sfig23}
\end{subfigure}
\begin{subfigure}{.5\textwidth}
  \centering
  \includegraphics[width=1\linewidth]{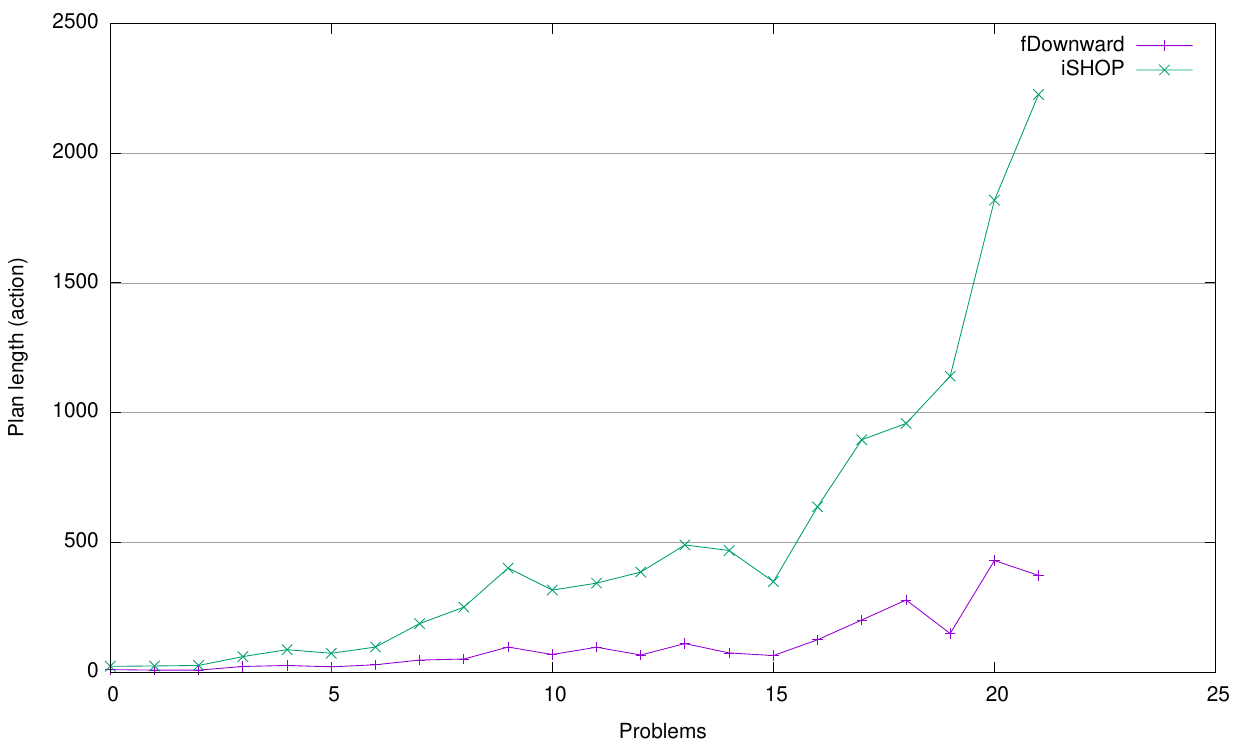}
  \caption{\scriptsize{Longueur des plans de f\_downward et iSHOP sur le domaine rover}}
  \label{fig:sfig24}
\end{subfigure}
\caption{Comparaison de la longueur de plans des algorithmes iSHOP, Fast Downward et SHOP sur les domaines de planification: rover, childsnack et satellite}
\label{fig:fig2}
\end{figure}
\end{center}

%
% Conclusion
%
\section{Conclusion}
Nous avons présenté, dans cet article, une nouvelle approche totalement instanciée pour la planification HTN. La particularité de cette approche réside dans le fait qu'elle réutilise les méthodes d'instanciation et de simplification utilisées dans la planification classique, et propose de nouvelles règles pour l'instanciation des méthodes HTN. Nous avons utilisé pour cela une représentation des méthodes ayant une grande expressivité permettant d'être utilisée par tous les types de planificateur HTN à la fois dans l'espace de plan et à la fois dans l'espace d'états. Nous avons cherché à démontrer l'efficacité de notre approche sur trois domaines de planification à travers l'implémentation et le test de l'algorithme iSHOP. Les résultats obtenus avec ce dernier montrent qu'une approche de planification HTN totalement instanciée permet d'avoir des temps de recherches beaucoup plus courts qu'avec une approche HTN classique. L'avantage de l'approche instanciée ne réside pas seulement dans le temps de traitement, puisqu'elle permet d'effectuer des études atteignabilité à partir des méthodes complètement instanciées et ouvre la porte à l'utilisation d'heuristiques de recherche.

Au vu des résultats obtenus lors de la première phase de tests, nous envisageons de continuer de faire plus de tests sur d'autres domaines de planification, afin d'avoir plus de points de comparaisons et valider les résultats actuels.
Nous envisageons aussi dans de futurs travaux de proposer et d'implémenter des heuristiques de recherche avec l'algorithme iSHOP et un autre algorithme HTN utilisant l'espace de plans qui est en cours de développement au sein de l'équipe et tester les performances de l'approche HTN totalement instanciée avec heuristiques sur les deux types d'algorithmes.

\newcommand{\etalchar}[1]{$^{#1}$}

\end{document}